\documentclass[final]{cvpr}
\PassOptionsToPackage{bookmarks=false}{hyperref}
\usepackage{times}
\usepackage{epsfig}
\usepackage{graphicx}
\usepackage{amsmath}
\usepackage{amssymb}
\usepackage{comment}
\usepackage{booktabs}
\usepackage{xcolor}
\usepackage{microtype}
\usepackage{enumitem}
\usepackage[symbol]{footmisc}

\usepackage[pagebackref=true,breaklinks=true,colorlinks,bookmarks=false]{hyperref}

\def\cvprPaperID{4028} 
\def\confYear{CVPR 2021}

\begin{document}

\title{SGCN:Sparse Graph Convolution Network for Pedestrian Trajectory Prediction}

\author{Liushuai Shi$^1$ ~Le Wang$^{2}$\footnote[1]{}  ~~Chengjiang Long$^3$ ~Sanping Zhou$^2$ ~Mo Zhou$^2$ ~Zhenxing Niu$^4$ ~Gang Hua$^5$ \\
$^{1}$School of Software Engineering, Xi'an Jiaotong University\\
$^{2}$Institute of Artificial Intelligence and Robotics, Xi'an Jiaotong University\\
$^{3}$JD Finance America Corporation, $^{4}$Machine Intelligence Lab, Alibaba Group $^{5}$Wormpex AI Research\\

}

\maketitle
\pagestyle{empty}  
\thispagestyle{empty} 
\footnotetext{$^*$Corresponding author.}
\begin{abstract}
Pedestrian trajectory prediction is a key technology in autopilot, which remains to be very challenging due to complex interactions between pedestrians.
However, previous works based on dense undirected interaction suffer from modeling superfluous interactions and neglect of trajectory motion tendency, and thus inevitably result in a considerable deviance from the reality.
To cope with these issues, we present a Sparse Graph Convolution Network~(SGCN) for pedestrian trajectory prediction.
Specifically, the SGCN explicitly models the sparse directed interaction with a sparse directed spatial graph to capture adaptive interaction pedestrians.
Meanwhile, we use a sparse directed temporal graph to model the motion tendency, thus to facilitate the prediction based on the observed direction.
Finally, parameters of a bi-Gaussian distribution for trajectory prediction are estimated by fusing the above two sparse graphs.
We evaluate our proposed method on the ETH and UCY datasets, and the experimental results show our method outperforms comparative state-of-the-art methods by $9\%$ in Average Displacement Error~(ADE) and $13\%$ in Final Displacement Error~(FDE).
Notably, visualizations indicate that our method can capture adaptive interactions between pedestrians and their effective motion tendencies.
\end{abstract}

\section{Introduction}


Given the observed trajectories of pedestrians, pedestrian trajectory prediction aims to predict a sequence of future location coordinates of pedestrians, which plays a critical role in various applications like autonomous driving~\cite{bai2015intention,luo2018porca}, video surveillance~\cite{luber2010people,yasuno2004pedestrian} and visual recognition~\cite{donahue2015long,Long:CVPR2017,Hu:TIP2021}.

Despite the recent advances in the literature, pedestrian trajectory prediction remains to be a very challenging task due to the complex interactions between pedestrians. For example, the motion of a pedestrian is very easy to be disturbed by other pedestrians~\cite{gupta2018social},  close friends or colleagues are likely to walk in groups~\cite{mohamed2020social}, and different pedestrians usually conduct similar social actions~\cite{sun2020recursive}.
To model the interactions between pedestrians, extensive works~\cite{mehran2009abnormal,alahi2014socially,gupta2018social,kosaraju2019social,ivanovic2019trajectron,mohamed2020social,10.1007/978-3-030-58610-2_30} have been done in the past few years, in which the weighting-by-distance methods~~\cite{mehran2009abnormal,alahi2014socially,gupta2018social,mohamed2020social} and the attention-based methods~\cite{kosaraju2019social,ivanovic2019trajectron,10.1007/978-3-030-58610-2_30, Ding:ICCV2019, Islam:CVPR2020, Islam:AAAI2021} have  achieved the state-of-the-art results in pedestrian trajectory prediction.

\begin{figure}[t]
\begin{center}
\includegraphics[width=\linewidth]{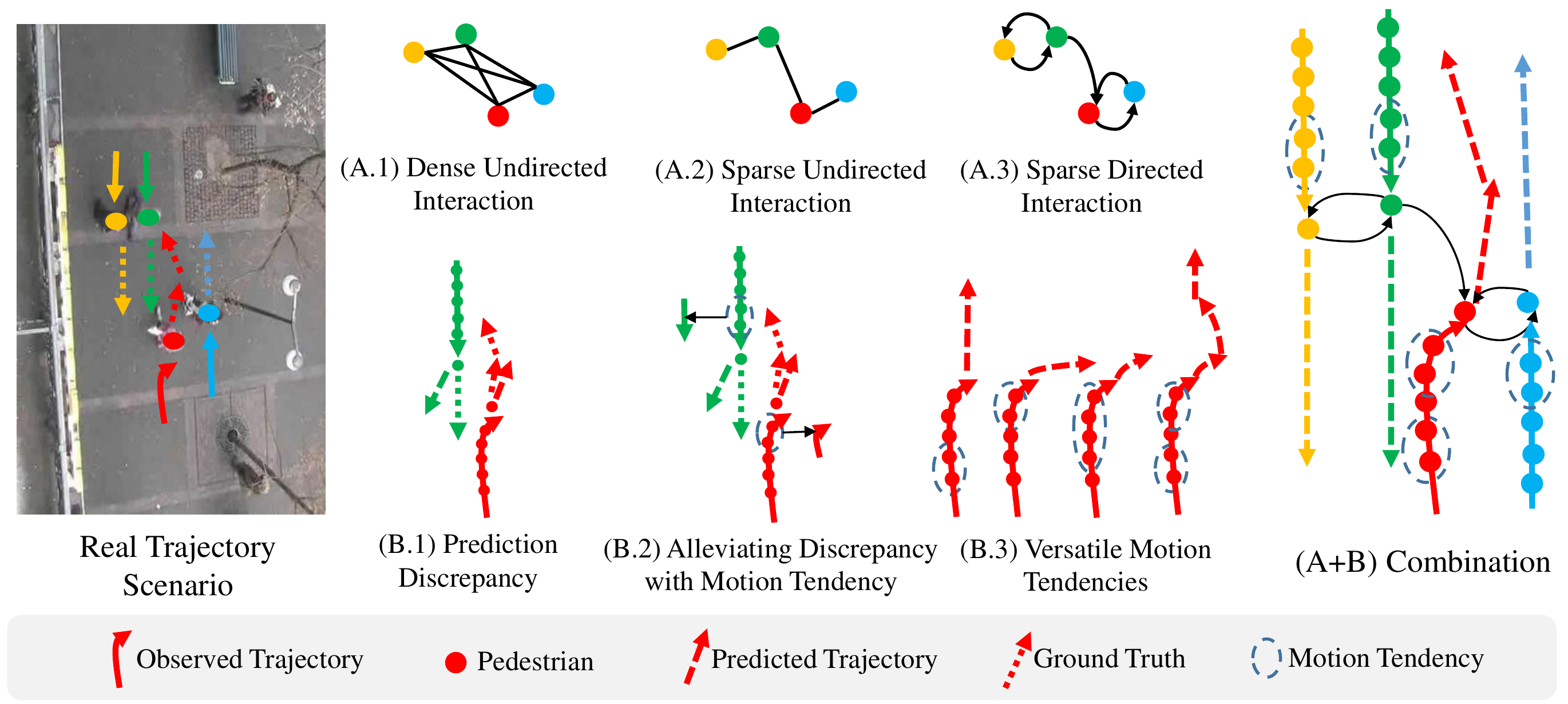}
\end{center}
   \caption{
Sparse Directed Interaction \& Motion Tendency.  Different pedestrians are marked in different colors.
\textbf{(A.1)} \emph{Dense undirected} interaction, where any pedestrian interacts with all other pedestrians.
\textbf{(A.2)} \emph{Sparse undirected} interaction with superfluous interactions being removed.
\textbf{(A.3)} \emph{Sparse directed} interaction with adaptive interaction pedestrians.
\textbf{(B.1)} The predicted trajectory severely deviates from the ground-truth as the pedestrians try to avoid collision against each other.
\textbf{(B.2)} Trajectory points enclosed by the blue dotted circle indicate a \emph{motion tendency} which may be leveraged for trajectory prediction.
\textbf{(B.3)} Variation of motion tendencies with different sets of trajectory points.
   }
\label{figure1}
\end{figure}

Most of the weighting-by-distance  and attention-based methods take a dense interaction model to represent the complex interactions between pedestrians, in which they assume that a pedestrian interacts with all the rest pedestrians.
Besides, the weighting-by-distance methods apply the relative distance to model the undirected interaction, in which the interaction between two pedestrians are identical to each other.
However, we argue that both the dense interaction and undirected interaction will introduce the superfluous interactions between pedestrians. As shown in Figure~\ref{figure1}: \textbf{(1)}  two pairs of pedestrians head towards from the opposite direction, while only the trajectory of red pedestrian detours to avoid the collision with green pedestrian; and \textbf{(2)} the trajectories of blue and yellow pedestrians not influence each other.
It is obvious that the dense or sparse undirected interaction based methods will fail to deal with the interactions in this case. For example, the \emph{dense undirected interaction}, as represented by~{\textbf{A.1}}, will generate superfluous interactions between yellow and blue pedestrians, due to the trajectories of yellow and blue pedestrians do not influence each other.
Besides, the \emph{sparse undirected interaction}, as denoted in~{\textbf{A.2}}, generates the superfluous interactions between the green and red pedestrians, because the red pedestrian detours to avoid collision with the green pedestrian, while the green pedestrian walks straight forward. To solve the above problems, it's better to design a \emph{Sparse Directed Interaction}, as shown in~\textbf{A.3}, which can interact with the adaptive pedestrians in the prediction of pedestrian trajectory.

What's worse, previous works focus on collision avoidance, which leads to the predicted trajectories tend to generate detour trajectories to avoid the collision for green and red pedestrians, as indicated in~\textbf{B.1}, while the green pedestrian deviates from the ground truth.
In this case, we propose \emph{motion tendency}, which is represented by a short-term trajectory enclosed by the blue dotted circle as shown in~\textbf{B.2}, the trajectory direction of the green pedestrian is straight forward, and that of the red pedestrian deflects to avoid the collision with the green pedestrian.
Based on the assumption that the direction of a trajectory will not change too abruptly, the motion tendency is beneficial to the prediction for green pedestrian.
It should be noted that the motion tendency is versatile, as shown in~\textbf{B.3}, in which the last one performs better than others, because it can jointly capture the ``straight forward''
and ``temporary deviation'' tendencies. Once the effective set of intermediate points can be found, the motion tendency will facilitate pedestrian trajectory prediction.

\begin{figure*}[t]
\vspace{-0.2cm}
\centering
\resizebox{1.0\textwidth}{!}{%
\includegraphics[width=1.0\textwidth]{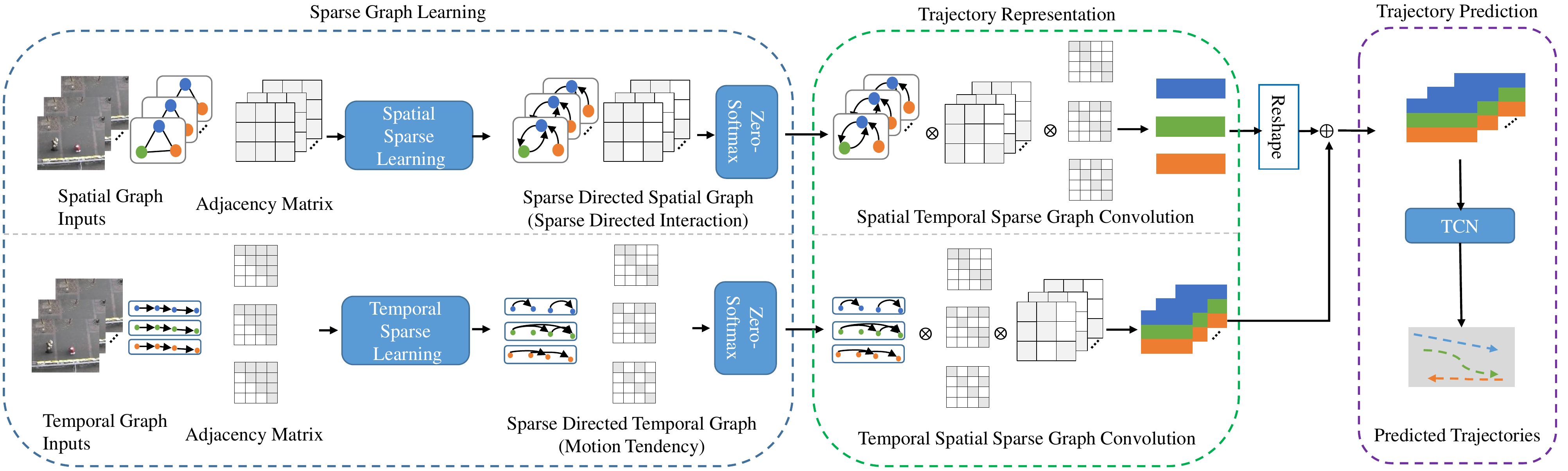}
}
\vspace{-0.2cm}
\caption{The framework of our proposed SGCN. The trajectories are reformed as spatial and temporal graph inputs.
Sparse Graph Learning involves the learning of sparse directed spatial
graph representing the Sparse Directed Interaction and sparse directed temporal graph representing
the Motion Tendency from the graph inputs.
Trajectory representations are learned by subsequent sparse spatial and
temporal graph convolution networks, and then fed into a TCN to estimate the
parameters of the bi-Gaussian distribution for future trajectory point
prediction.
}
\label{figure2}
\end{figure*}

\begin{figure*}[t]
\centering
\vspace{-0.5cm}
\resizebox{1.0\textwidth}{!}{
\includegraphics{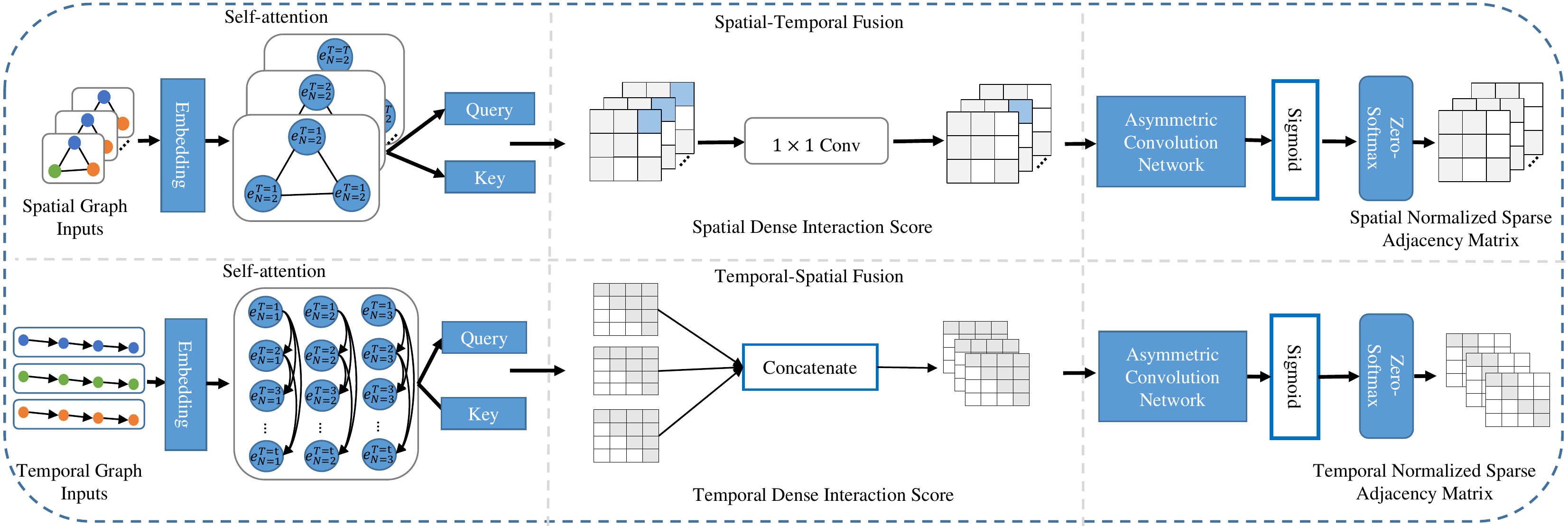}
}
\caption{Sparse Graph Learning.
The self-attention generates the \emph{dense} spatial interaction scores and
\emph{dense} temporal interaction scores based on the spatial and temporal
graph inputs, respectively.
Subsequent spatial-temporal fusion of the spatial interaction scores of each
time step and the temporal interaction scores of each pedestrian are done by $1\times1$ convolution layers
and self-attention mechanism.
The \emph{sparse} adjacency matrices are computed by asymmetric convolution
networks.
}
\label{figure3}
\end{figure*}

In this paper, we present a novel \emph{Sparse Graph Convolution Network}~(SGCN) which combines the Sparse Directed Interaction and Motion Tendency for pedestrian trajectory prediction. As shown in Figure~\ref{figure1}~(A+B), the Sparse Directed Interaction discover the set of pedestrians that effectively influence the trajectory of a particular pedestrian, and the  Motion Tendencies improve the future trajectory of interacted pedestrians.
In particular, as shown in Figure~\ref{figure2}, the Sparse Directed Spatial graph and Sparse Directed Temporal graph are jointly learned to model the Sparse Directed Interaction and the Motion Tendency of trajectory.
Specifically, the Sparse Graph Learning, as illustrated in Figure~\ref{figure3}, leverages self-attention~\cite{vaswani2017attention} mechanism to learn the asymmetric dense and directed interaction scores between trajectory points.
Then, these interaction scores are fused and fed into asymmetric convolutional networks to obtain high-level interaction features. Finally, a \emph{sparse directed spatial} and a \emph{sparse directed temporal} adjacency matrix can be obtained after pruning the superfluous interactions using a constant threshold and a normalization step of our ``Zero-Softmax'' function. The final asymmetric normalized sparse directed adjacency matrices can represent the sparse directed graph. 
 Once the above two graphs are obtained, we further learn the trajectory representation by a cascade of Graph Convolution Networks~\cite{kipf2016semi}, and employ the Time Convolution Network~\cite{bai2018empirical}  to estimate the parameters of the bi-Gaussian distribution, which are used to generate the predicted trajectories.
 %
 %

Extensive experimetal results on the
ETH~\cite{pellegrini2009you} and UCY~\cite{lerner2007crowds} datasets show that our method outperforms all the comparison state-of-the-art works.

To our best knowledge, this is the first work that explicitly models the Sparse Directed Interaction and Motion Tendency. In summary, our contributions are three-fold:~\textbf{(1)} we propose to model the \emph{Sparse Directed Interaction} and \emph{Motion Tendency} to improve the predicted trajectories; \textbf{(2)} we design an adaptive method to model the Sparse Directed Interaction and Motion Tendency; and \textbf{(3)} we propose
a sparse graph convolution network to learn the trajectory representations, where the advantage of explicit sparsity is demonstrated by the experiments.

\section{Related Works}

\textbf{Pedestrian Trajectory Prediction.}
Thanks to its powerful representational ability, deep learning becomes increasingly prevalent for predicting the pedestrian trajectories.
Social-LSTM~\cite{alahi2016social} models the trajectory of each pedestrian with Recurrent Neural Networks~(RNNs)~\cite{hochreiter1997long,jozefowicz2015empirical,chung2015gated}, and computes the interaction between pedestrians within a certain radius from the pooled hidden states.
SGAN~\cite{gupta2018social} predicts multi-modal trajectory using the Generative Adversarial Network~(GAN)~\cite{goodfellow2014generative,zhu2017unpaired,brock2018large},
and proposes a new pooling mechanism to compute interactions based on relative distance between pedestrians.
TPHT~\cite{ma2019trafficpredict} represents each pedestrian by an LSTM and employs
a soft-attention mechanism~\cite{vemula2018social} to model interactions between pedestrians.
Moreover, subsequent works leverage the scene features to improve the prediction accuracy.
PITF~\cite{liang2019peeking} considers the human-scene interaction and human-object interaction.
Sophine~\cite{sadeghian2019sophie} extracts scene features and social
features by a two-way attention mechanism, and computes the weights for all
agents with a social-attention.
TGFP~\cite{liang2020garden}  predicts both coarse and fine locations by using scene information.

Since the graph structure can better fit the scene, another track of works
model the human-human interaction using graph.
Social-BiGAT~\cite{kosaraju2019social} models the trajectory of each pedestrian
using LSTM, and the interactions by the Graph
Attention Network~(GAT)~\cite{velivckovic2017graph}.
To better represent the interaction between pedestrians,
Social-STGCNN~\cite{mohamed2020social} directly models the trajectory as a graph,
where the edges weighted by the pedestrian relative distance represent interactions between pedestrians.
RSGB~\cite{sun2020recursive} notes there are strong interactions between some
distant pedestrian pairs, hence invites sociologists to manually divide the
pedestrians into different groups according to specific physical rules and
sociological actions.
STAR~\cite{10.1007/978-3-030-58610-2_30} models the spatial interaction and temporal
dependencies by the Transformer~\cite{vaswani2017attention} framework.

In brief, previous works model the interactions for either the neighborhood
within a fixed physical range, or unexceptionally all pedestrians.
Presumably, this may result in discrepancies on the predictions due to superfluous
interactions.
In contrast, we propose a \emph{Sparse Directed Interaction}, which is capable of finding the adaptive pedestrians involved in the interaction, thus to alleviate such problem.
Besides, our method also captures the effective \emph{Motion Tendency}, which is helpful to improve the
accuracy of predicted trajectory. 

\textbf{Graph Convolution Networks.}
Graph convolution networks~(GCNs) are suitable for handling non-Euclidean data.
The existing GCN models can be divided into two categories:
1) the spectrum domain GCNs~\cite{kipf2016semi,defferrard2016convolutional} design the convolution operation based on Graph Fourier Transform.
It requires the adjacency matrix to be symmetric due to the eigen decomposition of Laplacian matrix;
2) the GCNs in spatial domain directly conduct convolution on the edge,
which is applicable on asymmetric adjacency matrices.
For example, GraphSage~\cite{hamilton2017inductive} aggregates the nodes in
three different ways and fuses adjacent nodes in different orders to extract node
features.
GAT~\cite{velivckovic2017graph} models the interaction between nodes using an
attention mechanism.
In order to deal with the spatio-temporal data,
STGCN~\cite{yan2018spatial} extends the spatial GCN to spatio-temporal
GCN for skeleton-based action recognition, which aggregates the nodes from a
local spatio-temporal scope.
Our SGCN differs from all the above GCNs, since it aggregates the nodes based
on a learned \emph{sparse} adjacency matrix, which means the set of nodes to be
aggregated is dynamically determined.

\textbf{Self-Attention Mechanism.}
The core idea of the Transformer~\cite{vaswani2017attention}, \ie{}, self-attention, has been demonstrated successfully in place of RNNs~\cite{jozefowicz2015empirical, chung2015gated} on a series of sequence modeling tasks in natural language processing,
such as text generation~\cite{yang2019xlnet}, machine translation~\cite{radford2018improving}, \emph{etc.}
Self-attention decouples the attention into the query, key and value which can
capture long-range dependencies, and takes advantage of parallel computation
compared with RNNs.
To represent the relationship between every pair of elements of the input
sequence, self-attention computes attention scores by a matrix multiplication
between the query and key.

In our method, we only compute a single layer attention scores to model
\emph{Sparse Directed Interaction} and \emph{Motion Tendency}.
Compared to the most recent work~\cite{10.1007/978-3-030-58610-2_30}, which predicts future trajectories by stacking Transformer block~(computation and memory expensive~\cite{hou2020dynabert}), our method is parameter-efficient and achieves better performance.
%


\section{Our Method}
Pedestrian trajectory prediction aims to predict future location coordinates of
pedestrians.
Given a series of observed video frames over time $t \in \{1,2,\ldots,T_\text{obs} \}$,
we can obtain the spatial~(2D-Cartesian) coordinates $\{(x_t^n, y_t^n)\}_{n=1}^N$ of all pedestrians with a tracking algorithm.
Based on these trajectories, our objective is to predict the pedestrian coordinates
within a future time $t \in \{T_{\text{obs}}+1, T_{\text{obs}}+2, \ldots,
T_\text{pred} \}$.

As discussed above, the existing works suffer from superfluous interactions by dense undirected graphs. Meanwhile, they also neglect the exploitable Motion Tendency clue.
To mitigate these limitations, we propose a Sparse Graph Convolutional Network (SGCN) for
trajectory prediction, which
mainly involves Sparse Graph Learning and bi-Gaussian distribution parameter
estimation based on the trajectory representations.
The overall architecture of the proposed network is represented in
Figure~\ref{figure2}.
First, the Sparse Directed Interaction~(SDI) and Motion Tendency~(MT) are learned from the
spatial and temporal graph inputs using self-attention mechanism and asymmetric
convolution networks, respectively.
Then, subsequent sparse spatial and temporal Graph Convolution Networks extract
the interaction and tendency features from the asymmetric adjacency matrices
representing sparse directed spatial graph (\emph{i.e.}, SDI) and sparse directed temporal
graph (\emph{i.e.}, MT).
Finally, the learned trajectory representations are fed into a Time
Convolution Network (TCN) to predict the parameters of a bi-Gaussian
distribution, which generates the predicted trajectory.

\subsection{Sparse Graph Learning}

\textbf{Graph Inputs.}
Given input trajectories $X_\text{in} \in \mathbb{R}^{T_\text{obs}\times N \times D }$, where $D$ denotes the dimension of spatial coordinate, we construct a spatial graph and a temporal graph as illustrated in Figure~\ref{figure3}.
The spatial graph $G_\text{spa} = (V^t, U^t)$ at time step $t$ represents locations of pedestrians, while temporal graph $G_\text{tmp}=(V_n, U_n)$ for pedestrian $n$ represents the corresponding trajectory.
$V^t=\{ v_n^t | n=1,\ldots,N \}$ and $V_n=\{ v_n^t | t=1,...,T_\text{obs} \}$ represent nodes of $G_\text{spa}$ and $G_\text{tmp}$, respectively, and the attribute of $v_n^t$ is the coordinate $(x_n^t, y_n^t)$ of the $n$-th pedestrian at time step $t$.
$U^t=\{ u_{i,j}^{t} | i,j=1,\ldots,N \}$ and $U_n=\{ u^{k,q}_{n} | k,q=1,\ldots,T_\text{obs} \}$ represent edges of $G_\text{spa}$ and $G_\text{tmp}$, respectively, where $u_{i,j}^{t},~u^{k,q}_{n} \in \{0,1\}$ indicate whether the nodes $v_i^t,v_j^t$ or nodes $v_n^k,v_n^q$ are connected~(denoted as 1) or disconnected~(denoted as 0), respectively.
Since there is no prior-knowledge on the connections of nodes, the elements in $U_n$ are initialized as 1, while $U^t$ is initialized as upper triangular matrix filled with $1$ because of the temporal dependency, namely the current state is independent to future states.

\textbf{Sparse Directed Spatial Graph.}
To increase the sparsity of the spatial graph inputs, \ie{}, identify the exact set of pedestrians involved in interactions in the spatial graph, we first adopt the self-attention mechanism~\cite{vaswani2017attention} to compute the asymmetric attention score matrix, namely the \emph{dense} spatial interaction $R_\text{spa}\in \mathbb{R}^{N\times N}$ between pedestrians, as follows:
\begin{equation} \label{SA}
\begin{split}
E_\text{spa} &=  \phi(G_\text{spa},W_E^\text{spa}),  \\
Q_\text{spa} &=  \phi(E_\text{spa},W_Q^\text{spa}),  \\
K_\text{spa} &=  \phi(E_\text{spa},W_K^\text{spa}),  \\
R_\text{spa} &=  \text{Softmax}(\frac{Q_\text{spa} K_\text{spa}^\texttt{T}}{\sqrt{d_\text{spa}}}),
\end{split}
\end{equation}
where $\phi(\cdot,\cdot)$ denotes linear transformation, $E_\text{spa}$ are the graph
embeddings, $Q_\text{spa}$ and $K_\text{spa}$ are the query and key of
the self-attention mechanism, respectively.
$W_E^\text{spa} \in \mathbb{R}^{D \times D_E^\text{spa}}$,
$W_Q^\text{spa} \in \mathbb{R}^{D \times D_Q^\text{spa}}$,
$W_K^\text{spa} \in \mathbb{R}^{D \times D_K^\text{spa}}$
are weights of the linear transformations, and
$\sqrt{d_\text{spa}}=\sqrt{D_Q^\text{spa}}$ is a scaled
factor~\cite{vaswani2017attention} to ensure numerical stability.

Since $R_\text{spa}$ is computed at every time step independently, it does not
contain any temporal dependency information of the trajectories.
Hence, we stack the \emph{dense} interactions $R_\text{spa}$ from every time
step as $R_\text{spa}^{\text{s-t}} \in \mathbb{R}^{T_{obs} \times N \times N}$,
and then fuse these stacked interactions with $1\times 1$ convolution along the
temporal channel, resulting in \emph{spatial-temporal dense} interactions
$\hat{R}_\text{spa}^{\text{s-t}} \in \mathbb{R}^{T_{obs} \times N \times N}$.

A slice of $\hat{R}_\text{spa}^{\text{s-t}}$ at each time step is an asymmetric
square matrix, where its $(i,j)$-th element represents the influence of node
$i$ to node $j$.
Then, the initiative and passive relations represented in the rows and columns
of the matrix respectively can be combined to obtain high-level interaction
features.
Specifically, a cascade of asymmetric convolution
kernels~\cite{szegedy2016rethinking} are applied on the rows and columns of
$\hat{R_\text{spa}^{\text{s-t}}}$, respectively,
%
\emph{i.e.},
\begin{equation}
\begin{split}
F_\text{row}^{(l)} &=\text{Conv}\left(F^{(l-1)}, \mathcal{K}^\text{row}_{(1\times S)} \right),  \\
F_\text{col}^{(l)} &=\text{Conv}\left(F^{(l-1)}, \mathcal{K}^\text{col}_{(S\times 1)} \right),  \\
F^{(l)} &=\delta\left(F_\text{row}^{(l)} +F_\text{col}^{(l)} \right),
\end{split}
\end{equation}
where $F_\text{row}^{(l)}$ and $F_\text{col}^{(l)}$ are the row-based and column-based asymmetric convolution feature maps at the $l$-th layer, respectively, $F^{(l)}$
is the activated feature map, and $\delta(\cdot)$ denotes a non-linear activation function.
$\mathcal{K}^\text{row}_{(1\times S)}$ and $\mathcal{K}^\text{col}_{(S\times 1)}$ are the convolution kernels of sizes $(1\times S)$ and $(S\times 1)$ (\emph{i.e.},~row and column vectors), respectively.
Note, $F^{(0)}$ is initialized as $\hat{R_\text{spa}^{\text{s-t}}}$, and all the convolution operations are padded with zeros in order to keep the output size as same as the input size.
Thus, the activated feature map obtained from the last convolution layer is the high-level interaction feature $F_\text{spa}$ of size $(T_{obs} \times N\times N)$.

We proceed to generate the sparse interaction mask $M_\text{spa}$ by element-wise threshold on $\sigma\left(F_\text{spa} \right)$  with a hyper-parameter $\xi\in [0,1]$.
When $F_\text{spa}[i,j] \geq \xi$, the $(i,j)$-th element of $M_\text{spa}$ is set to $1$, otherwise $0$, \emph{i.e.},
\begin{align}
\label{sgn}
M_\text{spa}=\mathbb{I}\left\{\sigma\left(F_\text{spa} \right) \geq \xi\right\},
\end{align}
where $\mathbb{I}\{\cdot\}$ is the indicator function, which outputs $1$ if the corresponding inequality holds, otherwise $0$. The $\sigma$ is Sigmoid activation function.
To ensure the nodes are self-connected, we add an identity matrix $I$ to the interaction mask, and then fuse it with the spatial-temporal dense interaction $\hat{R_\text{spa}^{\text{st}}}$ by element-wise multiplication, resulting in a sparse adjacency matrix $A_\text{spa}$, \emph{i.e.},
\begin{align}
A_\text{spa} &= (M_\text{spa}+I) \odot \hat{R_\text{spa}^{\text{s-t}}},
\end{align}
where $\odot$ denotes element-wise multiplication.

Some previous works (\emph{e.g.},~\cite{kipf2016semi}) suggest the normalization of
adjacency matrix is essential for GCN to function properly.
Nevertheless, the related works in the vertex domain directly adopt Softmax
function for adjacency matrix normalization, which leads to a side-effect that
the sparse matrix will be back to dense matrix because Softmax outputs non-zero values for zero inputs.
In this case, the pedestrians that do not interact with each other are forced
to interact with each other again.
To avoid this problem, we
design a ``Zero-Softmax`` function to
to keep the sparsity and the experimental results of ablation study represent the ``Zero-Softmax`` can further improve the performance.
Specifically, given a flattened matrix $\mathbf{x}=[x_1,x_2,\ldots,x_\mathcal{D}]$,
\begin{align}
    \text{Zero-Softmax}(x_i) &= \frac{(\exp(x_i)-1)^2}
    {\sum_j^\mathcal{D} (\exp(x_j)-1)^2 + \epsilon},
\end{align}
where $\epsilon$ is a neglectable small constant to ensure numerical stability,
and $\mathcal{D}$ is the dimension of the input vector.
Upon this, we can obtain the normalized sparse adjacency matrix $\hat{A}_{spa}
= \text{Zero-Softmax}(A_{spa})$.
Thus, a spatial-temporal sparse directed graph
$\hat{G}_\text{spa}=(V^t,\hat{A}_\text{spa})$ representing the Sparse Directed
Interactions is eventually obtained from the spatial graph inputs.
The whole process is illustrated in Figure~\ref{figure3}.

\textbf{Sparse Directed Temporal Graph.}
Following a similar way with the sparse directed spatial graph, we can also obtain the effective Motion Tendency,
namely the normalized adjacency matrix $\hat{A}_\text{tmp}$ from the temporal
graph inputs, except for two differences.

First, a position encoding tensor $\mathcal{E}$ \cite{vaswani2017attention} is
added to $E_\text{tmp}$,
\emph{i.e.}, $E_\text{tmp}=\phi(G_\text{tmp},W_E^\text{tmp}) + \mathcal{E}$,
because trajectory points in different order indicate different Motion
Tendencies.
Notably, the dense temporal interaction $R_\text{tmp}$ is also an upper
triangular matrix like $U^t$ due to temporal dependency.

The second difference lies in the temporal-spatial fusion step as illustrated in Figure~\ref{figure3}, where we can not perform convolution on $R_\text{tmp}^\text{t-s} \in \mathbb{R}^{N \times T_{obs} \times T_{obs}}$ obtained by stacking $R_\text{tmp} \in \mathbb{R}^{ T_{obs} \times T_{obs}}$, because the number of pedestrians $N$ is variable for different scenes. 
To simplify operation, we directly view the $R_\text{tmp}^\text{t-s}$ as the temporal-spatial dense interaction.
%
%

Thus, we eventually obtain a temporal-spatial
sparse directed graph $\hat{G}_\text{tmp}=(V_n,\hat{A}_\text{tmp})$
representing the Motion Tendency from the temporal graph inputs.

\begin{table*}[t]
\normalsize
\centering
\resizebox{1.0\linewidth}{!}{
\setlength{\tabcolsep}{1.0em}%
\begin{tabular}{cc|ccccc|c}
\toprule
Model   &   Year           & ETH & HOTEL & UNIV & ZARA1 & ZARA2 & AVG      \\
\midrule
Vanilla LSTM~\cite{alahi2016social} & 2016   & 1.09/2.41 & 0.86/1.91 & 0.61/1.31 & 0.41/0.88 & 0.52/1.11 & 0.70/1.52 \\
Social LSTM~\cite{alahi2016social} & 2016   & 1.09/2.35 & 0.79/1.76 & 0.67/1.40 & 0.47/1.00 & 0.56/1.17 & 0.72/1.54 \\
SGAN~\cite{gupta2018social} & 2018   & 0.87/1.62 & 0.67/1.37 & 0.76/1.52 & 0.35/0.68 & 0.42/0.84 & 0.61/1.21 \\
Sophie~\cite{sadeghian2019sophie}& 2019    & 0.70/1.43 & 0.76/1.67 & 0.54/1.24 & 0.30/0.63 & 0.38/0.78 & 0.51/1.15 \\
PITF~\cite{liang2019peeking} & 2019   & 0.73/1.65 & 0.30/0.59 & 0.60/1.27 & 0.38/0.81 & 0.31/0.68 & 0.46/1.00 \\
GAT~\cite{kosaraju2019social} & 2019  & 0.68/1.29 & 0.68/1.40 & 0.57/1.29 & \textbf{0.29}/0.60 & 0.37/0.75 & 0.52/1.07 \\
Social-BIGAT~\cite{kosaraju2019social} & 2019 & 0.69/1.29 & 0.49/1.01 & 0.55/1.32 & 0.30/0.62 & 0.36/0.75 & 0.48/1.00 \\
Social-STGCNN~\cite{mohamed2020social} & 2020 & 0.64/1.11 & 0.49/0.85 & 0.44/0.79 & 0.34/\textbf{0.53} & 0.30/0.48 & 0.44/0.75 \\
RSBG w/o context~\cite{sun2020recursive} & 2020 & 0.80/1.53 & 0.33/0.64 & 0.59/1.25 & 0.40/0.86 & 0.30/0.65 & 0.48/0.99 \\
STAR~\cite{10.1007/978-3-030-58610-2_30} & 2020 & \textbf{0.56}/1.11 & \textbf{0.26}/\textbf{0.50} & 0.52/1.15 & 0.41/0.90 & 0.31/0.71 & 0.41/0.87 \\
\hline
SGCN (Ours)       & -        & 0.63/\textbf{1.03} & 0.32/0.55 & \textbf{0.37}/\textbf{0.70} & \textbf{0.29}/\textbf{0.53} & \textbf{0.25}/\textbf{0.45} & \textbf{0.37}/\textbf{0.65} \\
\bottomrule
\end{tabular}
}
\caption{Comparison with the baselines approach on the public benchmark dataset ETH and UCY for ADE/FDE. All approaches input 8 frames and output 12 frames. Our SGCN significantly outperform the comparison state-of-the-art works. The lower the better. }
\vspace{-0.2cm}
\label{tab:result}
\end{table*}

\subsection{Trajectory Representation and Prediction}

GCNs can aggregate the nodes of sparse
graphs representing $\hat{A}_\text{spa}$ (SDI) and $\hat{A}_\text{tmp}$ (MT),
and learn the trajectory representation.
As illustrated in Figure~\ref{figure2}, we use two GCNs to learn the trajectory representation, where in one branch $\hat{A}_\text{spa}$ is fed to the network ahead
of $\hat{A}_\text{tmp}$, while in the other branch they are fed in the reverse
order.
Thus, the first branch produces interaction-tendency feature
$H_{\text{ITF}}$, while the other branch produces tendency-interaction feature
$H_{\text{TIF}}$, \emph{i.e.},
\begin{equation} \label{GCN}
\begin{split}
H^{(l)}_\text{ITF}&=\delta\left(\hat{A}_\text{tmp} \cdot  \delta(\hat{A}_\text{spa} H^{(l-1)}_{\text{ITF}} \mathbf{W}_\text{spa1}^{(l)}) \mathbf{W}_\text{tmp1}^{(l)}\right), \\
H^{(l)}_\text{TIF}&=\delta\left(\hat{A}_\text{spa} \cdot \delta(\hat{A}_\text{tmp} H^{(l-1)}_{\text{TIF}} \mathbf{W}_\text{tmp2}^{(l)}) \mathbf{W}_\text{spa2}^{(l)}\right),
\end{split}
\end{equation}
where $\mathbf{W}_\text{tmp1}$,$\mathbf{W}_\text{spa1}$,
$\mathbf{W}_\text{tmp2}$ and $\mathbf{W}_\text{spa2}$ are GCN weights, and $l$ represents the $l$-th layer of GCN.
$H^{(0)}_{\text{ITF}}$ is initialized as $\hat{G}_\text{spa}$, and
$H^{(0)}_{\text{TIF}}$ is initialized as $\hat{G}_\text{tmp}$.
The trajectory representation $H$ is the sum of the last GCN outputs
$H_{\text{ITF}}$ and $H_{\text{TIF}}$.

\textbf{Trajectory Prediction and Loss Function.}
We follow Social-LSTM~\cite{alahi2016social} to assume that the trajectory coordinates
$(x^{t}_{n},y^{t}_{n})$ at time step $t$ of pedestrian $n$ follow a
bi-variate Gaussian distribution
$\mathcal{N}\left(\hat{\mu}_{n}^{t}, \hat{\sigma}_{n}^{t},
\hat{\rho}_{n}^{t}\right)$,
where $\hat{\mu}_{n}^{t}$ is the mean, $\hat{\sigma}_{n}^{t}$ is the standard
deviation, and $\hat{\rho}_{n}^{t}$ is the correlation coefficient.
Given the final trajectory representation $H$, we can predict the parameters of
the bi-Gaussian distribution with a TCN~\cite{bai2018empirical} on the time dimension following Social-STGCNN~\cite{mohamed2020social}.
Note, TCN is chosen because it does not suffer from gradient vanishing and high
computational cost like traditional RNNs~\cite{hochreiter1997long,jozefowicz2015empirical,chung2015gated}.
Hence, the method can be trained by minimizing the negative log-likelihood loss as
\begin{align}
L^{n}(\mathbf{W}) &= -\sum_{t=T_\text{obs}+1}^{T_\text{pred}}
\log P\Big( (x_{n}^{t},y_{n}^{t}) ~\big|~ \hat{\mu}_{n}^{t}, \hat{\sigma}_{n}^{t}, \hat{\rho}_{n}^{t} \Big),
\end{align}
where $\mathbf{W}$ denotes all trainable parameters in the method.

\section{Experiments and Analysis}

\textbf{Evaluation Datasets.}  To validate the efficacy of our proposed method, we use two public pedestrian trajectory datasets, \emph{i.e.},
ETH~\cite{pellegrini2009you} and UCY~\cite{lerner2007crowds}, which are the most widely used benchmarks for the trajectory prediction task.
In particular, ETH dataset contains the ETH and HOTEL scenes,
while the UCY dataset contains three different scenes including UNIV, ZARA1, and ZARA2.
We use the ``leave-one-out''~\cite{sun2020recursive} method for training and evaluation.
We follow existing works that observing 8 frames~(3.2 seconds) trajectories and predicting the next 12 frames~(4.8 seconds).

\textbf{Evaluation Metrics.} We employ two metrics, namely Average Displacement Error (ADE)~\cite{raksincharoensak2016motion} and Final Displacement Error (FDE)~\cite{alahi2016social}
to evaluate the prediction result.
ADE measures the average L-2 distance between all the predicted trajectory points obtained from the method and all ground-truth future trajectory points,
while FDE measures the L-2 distance between the final predicted destination obtained from the method and final destination of the ground-truth future trajectory point.

\textbf{Experimental Settings.}
In our experiments, the embedding dimension of self-attention and the dimension of graph embedding are both set to $64$.
The number of self-attention layer is $1$.
The asymmetric convolution network comprises $7$ convolution layers with kernel size $S=3$.
The spatial-temporal GCN and temporal-spatial GCN cascade 1 layer, respectively. And the TCN cascade 4 layers.
The threshold value $\xi$ is empirically set to 0.5.
PRelu~\cite{7410480} is adopted as the nonlinear activation $\delta(\cdot)$.
The proposed method is trained using the Adam~\cite{kingma2014adam} optimizer for $150$ epochs with data batches of size $128$.
The initial learning rate is set to $0.001$, which is decayed by a factor $0.1$ with an interval of $50$ epochs.
During the inference phase, $20$ samples are drawn from the learned bi-variate Gaussian distribution and the closest sample to ground-truth is used to compute the ADE and FDE metrics.
Our method is implemented on PyTorch~\cite{NEURIPS2019_9015}.  
The code has been published\footnote[2]{code available at \url{https://github.com/shuaishiliu/SGCN}}.
\begin{table*}[t]
\normalsize
\centering
\resizebox{0.75\textwidth}{!}{
\begin{tabular}{c|ccccc|c}
\toprule
Variants                 & ETH & HOTEL & UNIV & ZARA1 & ZARA2 & AVG      \\
\midrule
\text{SGCN w/o MT }  & 0.92/1.23 & 0.69/1.53 & 0.61/1.80 & 0.52/0.60 & 0.40/0.80 & 0.62/1.19 \\
\text{SGCN w/o ZS}     & 0.73/1.39 & 0.34/0.59 & 0.38/0.75 & 0.34/0.65 & 0.26/\textbf{0.45} & 0.41/0.76 \\
\text{SGCN w/o SDI}     & 0.81/1.66 & 0.67/1.42 & 0.79/1.78 & 0.59/0.72 & 0.44/0.82 & 0.66/1.28 \\
\hline
SGCN (Ours)              &\textbf{0.63}/\textbf{1.03} & \textbf{0.32}/0.55 & \textbf{0.37}/\textbf{0.70} & \textbf{0.29}/0.53 & \textbf{0.25}/\textbf{0.45} & \textbf{0.37}/\textbf{0.65} \\
\bottomrule
\end{tabular}
}
\caption{The ablation study of each components. SGCN~(Ours) combines with each components.
}
\label{tab:ablation1}
\end{table*}

\begin{table*}[t]
\vspace{-0.2cm}
\normalsize
\centering
\resizebox{0.75\textwidth}{!}{
\begin{tabular}{c|ccccc|c}
\toprule
Variants                 & ETH & HOTEL & UNIV & ZARA1 & ZARA2 & AVG      \\
\midrule
\text{$\text{SGCN-V}_1$}     & 0.91/1.82 & 0.36/0.62 & 0.41/0.83 & 0.43/0.83 & 0.34/0.65 & 0.49/0.95 \\
\text{$\text{SGCN-V}_2$}     & 0.69/1.11 & \textbf{0.32}/0.57 & 0.41/0.78 & 0.31/0.53 & 0.27/\textbf{0.45} & 0.40/0.68 \\
\text{$\text{SGCN-V}_3$}     & 0.66/1.07 & 0.38/\textbf{0.46} & 0.54/0.77 & 0.30/\textbf{0.52} & \textbf{0.25}/0.47 & 0.42/\textbf{0.65} \\
\text{$\text{SGCN-V}_4$}     & 0.66/1.16 & 0.38/0.58 & 0.58/0.79 & 0.40/0.47 & 0.27/0.51 & 0.45/0.70 \\
\hline
SGCN (Ours)              &\textbf{0.63}/\textbf{1.03} & \textbf{0.32}/0.55 & \textbf{0.37}/\textbf{0.70} & \textbf{0.29}/0.53 & \textbf{0.25}/\textbf{0.45} & \textbf{0.37}/\textbf{0.65} \\
\bottomrule
\end{tabular}
}
\caption{The ablation study of threshold $\xi$. SGCN~(Ours) sets the $\xi=0.5$.
\vspace{-0.8cm}
}
\label{tab:ablation2}
\end{table*}

\subsection{Comparison with State-of-the-Arts}

We compare our method with nine state-of-the-art methods, including Vanilla LSTM~\cite{alahi2016social}, Social-LSTM~\cite{alahi2016social}, SGAN~\cite{gupta2018social}, Sophie~\cite{sadeghian2019sophie}, PITF~\cite{liang2019peeking}, Social-BiGAT~\cite{kosaraju2019social}, Social-STGCNN~\cite{mohamed2020social}, RSGB~\cite{sun2020recursive}, and STAR~\cite{yu2020spatio}, in the past four years.
The results are shown in Table~\ref{tab:result}, which are evaluated by using the ADE and FDE metrics.
The results indicate that our method significantly outperforms all
the competing methods on both the ETH and UCY datasets.
Especially for the ADE metric, our method surpasses the previous best method STAR~\cite{yu2020spatio} by 9\% averaging on ETH and UCY datasets.
For the FDE metric, our method is better than the previous best method Social-STGCNN~\cite{mohamed2020social} by a margin of 13\% averaging on the ETH and UCY datasets.
To our best knowledge, the under-lying reason is that our method can remove the interference from the superfluous interactions by leveraging Sparse Directed Interaction, and the Motion Tendency is leveraged to improve the prediction.

Interestingly, our method outperforms all the dense interaction based methods, such as SGAN~\cite{gupta2018social}, Sophie~\cite{sadeghian2019sophie}, GAT~\cite{kosaraju2019social}, Social-BiGAT~\cite{kosaraju2019social}, Social-STGCNN~\cite{mohamed2020social}, and STAR~\cite{yu2020spatio}, on the UNIV sequence which mainly contains dense crowd scenes. We speculate that the dense interaction based methods may capture superfluous interaction objects, which will cause prediction discrepancies. What's different, our method is capable of removing the superfluous interactions by Sparse Directed Interaction, which is benefit to obtain a better performance.

\subsection{Ablation Study}
Firstly, we conduct ablative experiments on both ETH and UCY datasets, so as to isolate the contribution of each component to the final performance. Secondly, we set different values of threshold $\xi$ to evaluate the effectiveness of the proposed sparse graph with different sparsity. The detailed experiments are introduced in the following paragraphs.

\textbf{Contribution of Each Component.} As illustrated in Table~\ref{tab:ablation1}, we evaluate three different variants of our method, in which: \textbf{(1)}~{SGCN w/o MT} denotes the Motion Tendency is removed in our method, in which it merely models the Sparse Directed Interaction;
\textbf{(2)}~{SGCN w/o ZS} indicates that the Zero-Softmax is replaced by Softmax for sparse adjacency matrix normalization; and \textbf{(3)}~{SGCN w/o SDI} represents that the Sparse Directed Interaction is  removed in our method, in which it merely models the Motion Tendency.
From the results, we can see that removing any component from our model will result in a large performance reduction.
In particular, the results of SGCN w/o MT show $67\%$ performance degradation in ADE and $83\%$ in FDE,  which clearly validate the contribution of the Motion Tendency to the final performance of pedestrian trajectory prediction.
Besides, the results of SGCN w/o SDI show $78\%$ performance degradation in ADE and $96\%$ in FDE, which indicate that the sparse directed interaction is also important for the pedestrian trajectory prediction .

\textbf{Effectiveness of Sparse Graph.}
As illustrated in Table~\ref{tab:ablation2}, we evaluate five different variants of our method, in which:
\textbf{(1)}~{$\text{SGCN-V}_1$}: it means there is no interaction between each pair of pedestrians by setting $\xi=1$;
\textbf{(2)}~{$\text{SGCN-V}_2$}: it leads to very sparse directed interaction by setting $\xi=0.75$;
\textbf{(3)}~{$\text{SGCN-V}_3$}: it leads to relatively dense directed interaction by setting $\xi=0.25$;
\textbf{(4)}~{$\text{SGCN-V}_4$}: it leads to dense interactions by setting $\xi=0$; and
\textbf{(5)}~{\text{SGCN}}: it responds to our full method by setting $\xi=0.5$.
The experimental results are presented in Table~\ref{tab:ablation2}.
%
%
%
We find that the overall performances of our method reaches a peak when $\xi=0.5$,
which means enforcing sparsity to a certain extent is effective enough.
Besides, $\text{SGCN-V}_1$ achieves the lowest performance, implying the necessity of modeling interactions between pedestrians.
Furthermore, the results of $\text{SGCN-V}_2$ and $\text{SGCN-V}_3$ are better than that of $\text{SGCN-V}_4$, which indicates that the sparse interaction indeed can lead to performance improvement.

\subsection{Visualization}

\begin{figure*}[t]
\vspace{-0.4cm}
\centering
\resizebox{0.90\textwidth}{!}{
\includegraphics{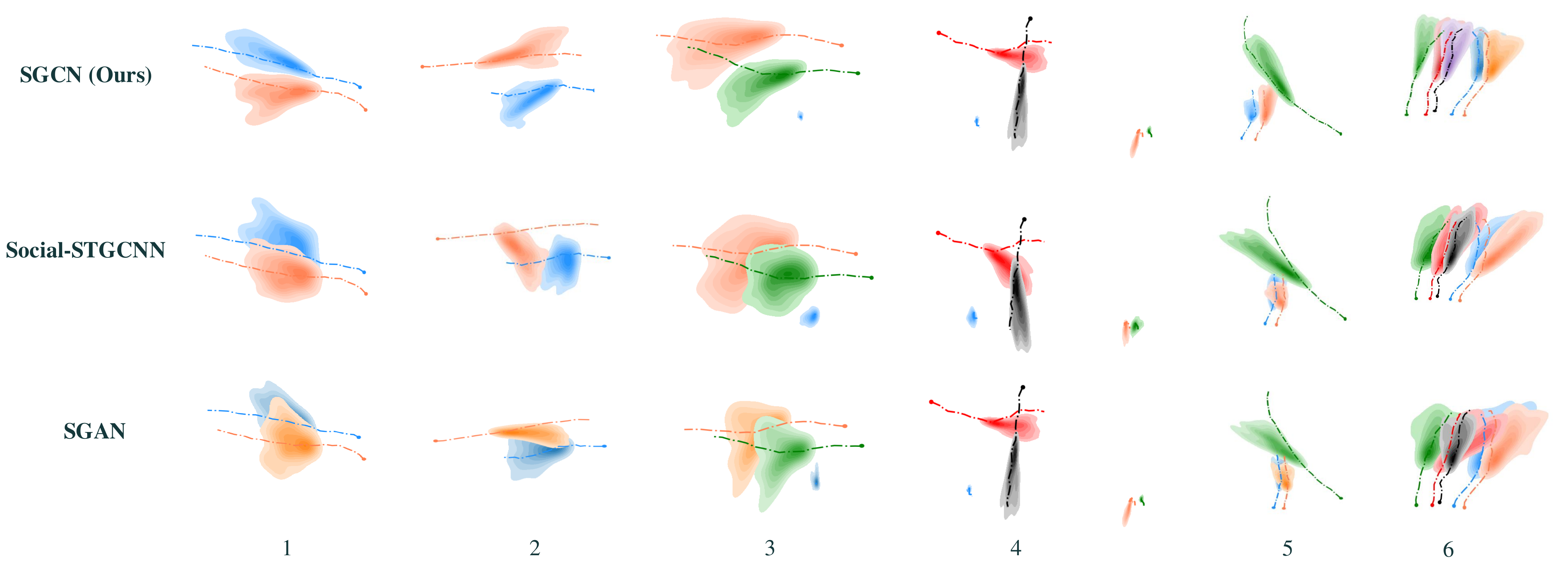} 
}
\caption{\textbf{Visualization of predicted trajectories distribution.} 
Different colors represents different pedestrians. 1 and 2 shows two pedestrians walking in parallel from the same direction and different direction, respectively. 3 and 4 shows the scene where two pedestrians meet. 5 shows a pedestrian meets multiple pedestrians. 6 shows several pedestrians walking side-by-side. }
\label{figure4}
\end{figure*}

\begin{figure*}[t]
\centering
\vspace{-0.4cm}
\resizebox{0.90\textwidth}{!}{
\includegraphics[width=\textwidth]{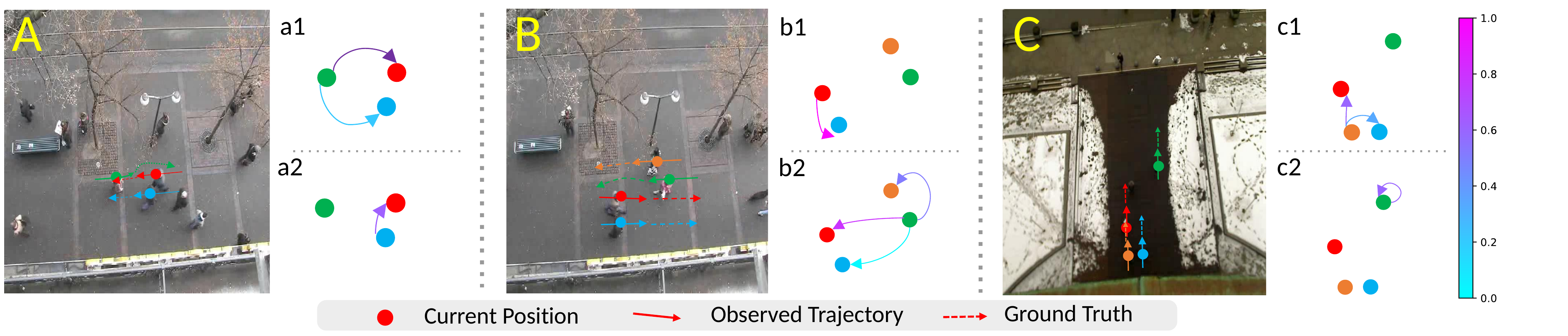} 
}
\caption{\textbf{Visualization of Sparse Directed Interaction.} The rightmost colorbar indicates the weight of SDI corresponding to different colors. In particular, purple indicates strong relationship while blue means week relationship. Some pedestrians are unmarked because there is no record in the dataset.}
\vspace{-0.20cm}
\label{figure5}
\end{figure*}

\textbf{Trajectory Prediction Visualization.}
%
%
We 
visualize several common interaction scenes in Figure~\ref{figure4}, where the solid dot at the end of each trajectory
denotes the start. More scene visualizations will be presented in the supplementary materials.
We compare our method with Social-STGCNN~\cite{mohamed2020social} and SGAN~\cite{gupta2018social}, because both of them learn a parameterized distribution of future trajectories.
%

Specifically, the scenarios $1$ and $2$ depict two pedestrians walking in parallel in the same or opposite direction, respectively.
In these cases, the pedestrians are not likely to collide.
The visualization reveals that our predicted distribution has a better tendency along the ground-truth, while both the Social-GCNN and SGAN generate larger overlap which implies potential collisions, and thus deviate from the ground-truth.
The scenarios $3$ and $4$ show two pedestrians heading towards another one that stays still, and one pedestrian meets another pedestrian in a perpendicular direction, respectively. The Social-STGCNN and SGAN again both suffer from the overlap issue, indicating high possibility of collision, while there are less overlaps in our predicted distribution. Particularly,  the green pedestrian stands still in scenario $3$, thus our predicted distribution has smaller variance, indicating our method captures the fact that the still pedestrian is not influenced by other pedestrians in scenario $3$.
The scenarios $5$ and $6$ represent the meeting of more than one pedestrian, where our results considerably match the ground-truth, while the results of Social-STGCNN and SGAN have serious overlap and diverge from the ground-truth.

To summarize, both Social-STGCNN and SGAN predict overlapping distributions and deviate from the ground-truth, while our predicted distributions exhibit less overlapping and have a better tendency along the ground-truth.
For the overlaps, the reason maybe that Social-STGCNN and SGAN model the dense interaction which inevitably introduces superfluous interactions to disturb the normal trajectory and generate a large detour to avoid collision.
%
%
%
%
In contrast, SGCN models the sparse directed interaction and motion tendency together and leads to a better prediction distribution.

%
%
%
%
%


\textbf{Sparse Directed Interaction Visualization.}
The Sparse Directed Interaction is visualized in Figure~\ref{figure5}, from which we find that our method is able to capture effective interaction objects on different interaction scenes. %
The graphs (a2), (b1), (c1) and (c2) illustrate the sparse directed interactions that one node is only influenced by part of other nodes. For instance, the graph (a2) represents  the sparse directed interaction between the blue node and red node, and it conforms the scene of A, where the trajectory of blue node is only influenced by the red node according to the ground-truth.
Furthermore, we find our method can capture interaction objects dynamically, except the sparse directed interaction given by graphs (a2), (b1), (c1) and (c2).  The graphs (a1) and (b2) show the green node interacts with all marked nodes.

\section{Conclusion}
In this paper, we present a  sparse graph convolution network for trajectory prediction, which leverages the Sparse Directed Interaction and Motion Tendency.
According to the extensive experimental evaluations, our method achieves better performances than previous methods.
Moreover, our method can predict trajectories more accurately even under some complex scenes, such as a group of pedestrians walking in parallel.
These improvements can be attributed to the abilities of identifying the Sparse Directed Interactions and Motion Tendencies of our method.

\section{Acknowledgment}

This work was supported partly by National Key R\&D Program of China Grant 2018AAA0101400, NSFC Grants 62088102, 61773312, and 61976171, Young Elite Scientists Sponsorship Program by CAST Grant 2018QNRC001, and Natural Science Foundation of Shaanxi Grant 2020JQ-069.

{\small
\bibliographystyle{ieee_fullname}
\bibliography{egbib}
}

\end{document}